\documentclass[letterpaper, 10 pt, conference]{ieeeconf}  

\IEEEoverridecommandlockouts                              

\usepackage{amsmath} 
\usepackage{amssymb}
\usepackage[mathscr]{euscript}
\usepackage{graphicx}
\usepackage[table]{xcolor}
\usepackage{algorithm}
\usepackage[noend]{algpseudocode}
\usepackage{tabularx}
\usepackage{multirow}
\usepackage{hhline}
\usepackage{hyperref}
\usepackage[T1]{fontenc}
\usepackage{lipsum}
\usepackage{stfloats}
\usepackage{subcaption}
\usepackage{lipsum}  
\usepackage{arydshln}
\usepackage{booktabs}
\usepackage{textcomp, gensymb}

\usepackage[capitalise,nameinlink]{cleveref}
  \crefname{section}{Sect.}{Sect.}
  \Crefname{section}{Section}{Sections}
  \crefname{figure}{Fig.}{Fig.}
  \Crefname{figure}{Figure}{Figures}
  \crefname{table}{Tabl.}{Tabl.}
  \Crefname{table}{Table}{Tables}

\usepackage{tikz}

\newcommand\copyrighttext{%
  \footnotesize \textcopyright \the\year{} IEEE. Personal use of this material is permitted. Permission from IEEE must be obtained for all other uses, including reprinting/republishing this material for advertising or promotional purposes, collecting new collected works for resale or redistribution to servers or lists, or reuse of any copyrighted component of this work in other works.}

\newcommand\copyrightnotice{%
\begin{tikzpicture}[remember picture,overlay]
\node[anchor=south,yshift=10pt] at (current page.south) {\fbox{\parbox{\dimexpr0.75\textwidth-\fboxsep-\fboxrule\relax}{\copyrighttext}}};
\end{tikzpicture}%
}

\title{\LARGE \bf
Learning Variable Compliance Control From a Few Demonstrations \\ for Bimanual Robot with Haptic Feedback Teleoperation System
}

\author{Tatsuya Kamijo$^{1, 2^*}$, Cristian C. Beltran-Hernandez$^{1^*}$, Masashi Hamaya$^{1}$
\thanks{$^{*}$ Equal contribution.}
\thanks{$^{1}$OMRON SINIC X Corporation, Tokyo, Japan}
\thanks{$^{2}$Department of Mechanical Engineering, The University of Tokyo, Japan}
\thanks{Corresponding authors:
        {\tt\small cristian.beltran@sinicx.com}, 
        {\tt\small tatsukamijo@ieee.org}%
        }%
}

\begin{document}
\setlength\textfloatsep{10pt}
\setlength\floatsep{0pt}



\maketitle
\thispagestyle{empty}
\pagestyle{empty}
\copyrightnotice

\begin{abstract}

Automating dexterous, contact-rich manipulation tasks using rigid robots is a significant challenge in robotics. Rigid robots, defined by their actuation through position commands, face issues of excessive contact forces due to their inability to adapt to contact with the environment, potentially causing damage.  While compliance control schemes have been introduced to mitigate these issues by controlling forces via external sensors, they are hampered by the need for fine-tuning task-specific controller parameters. Learning from Demonstrations (LfD) offers an intuitive alternative, allowing robots to learn manipulations through observed actions.
In this work, we introduce a novel system to enhance the teaching of dexterous, contact-rich manipulations to rigid robots. Our system is twofold: firstly, it incorporates a teleoperation interface utilizing Virtual Reality (VR) controllers, designed to provide an intuitive and cost-effective method for task demonstration with haptic feedback. Secondly, we present Comp-ACT (Compliance Control via Action Chunking with Transformers), a method that leverages the demonstrations to learn variable compliance control from a few demonstrations. Our methods have been validated across various complex contact-rich manipulation tasks using single-arm and bimanual robot setups in simulated and real-world environments, demonstrating the effectiveness of our system in teaching robots dexterous manipulations with enhanced adaptability and safety.
Code available at \hyperlink{https://github.com/omron-sinicx/CompACT}{https://github.com/omron-sinicx/CompACT}.

\end{abstract}

\section{Introduction}

Dexterous robotic manipulation holds significant importance in various fields due to its ability to automate tasks that involve physical interaction with objects. Here, we consider widely used rigid robots, such as collaborative or industrial robots, that can only be actuated through position/velocity commands. 
We adopt compliance control schemes that allow rigid robots to handle tasks involving direct contact by controlling contact forces through an external sensor~\cite{calanca2015review}. 
However, programming robots to do dexterous contact-rich manipulations remains an open challenge~\cite{kroemer2021review}.


Learning from Demonstrations (LfD) has emerged as an intuitive and promising solution to teach robots dexterous manipulations
~\cite{zhu2018robot}.
Instead of manually programming a sequence of actions or engineering a reward function for the robot to learn from, in LfD, the operator can show the robot how exactly to perform the task. However, the challenges of effectively using LfD are sample efficiency-the need for many demonstrations for each task-, selecting optimal stiffness parameters for the compliance controller, and the teaching interface. We propose a robot demonstration system for contact-rich manipulation tasks to address these challenges.

\begin{figure}[t]
    \centering
    \includegraphics[width=0.95\linewidth]{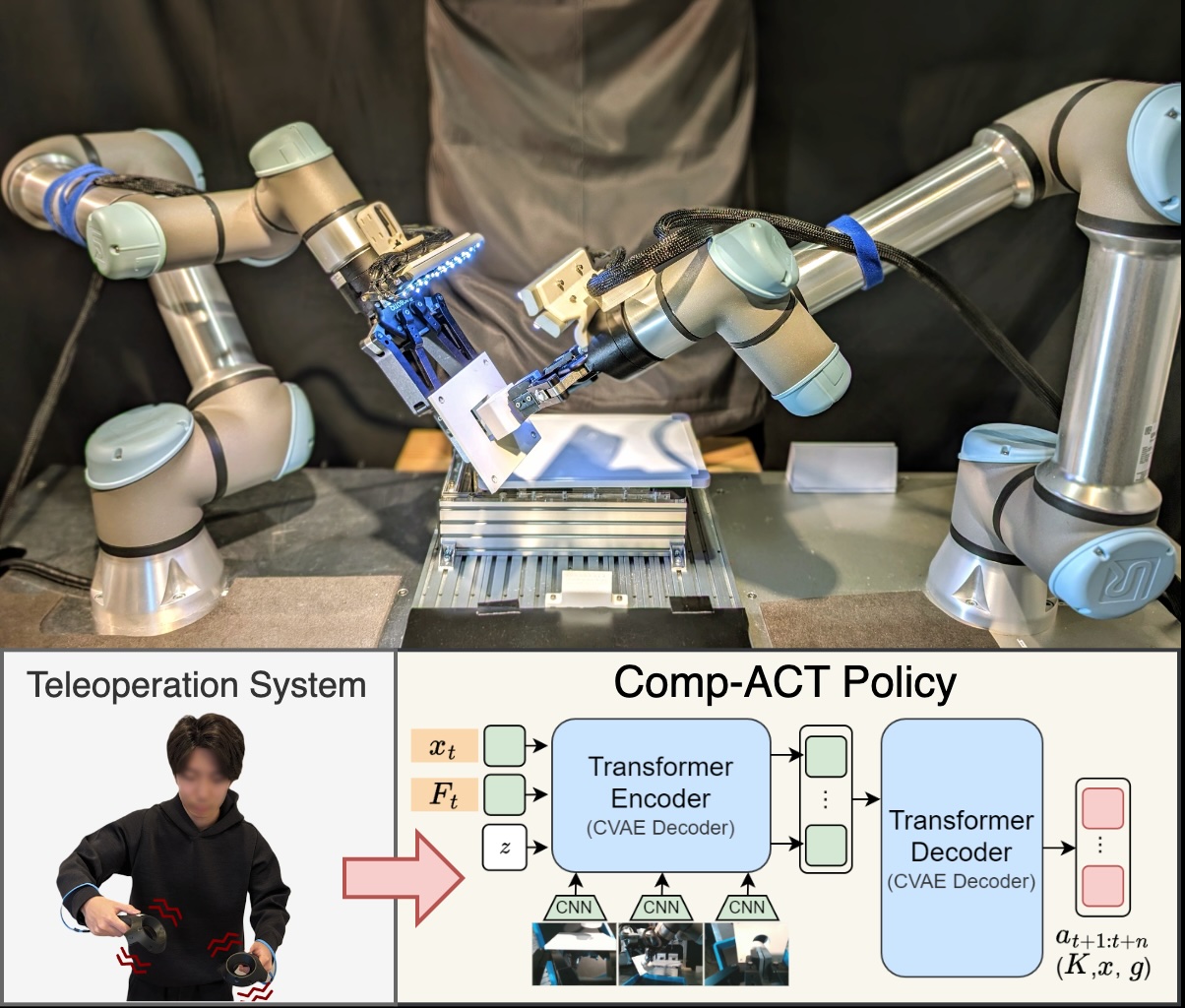}
    \caption{Proposed teleoperation interface via VR controllers with haptic feedback through vibrations (\textit{bottom left}) and LfD method Comp-ACT: \textit{\underline{Comp}liance Control via \underline{A}ction \underline{C}hunking with \underline{T}ransformers} (\textit{bottom right}). Our system was evaluated on complex contact-rich tasks on a dual-arm robotic setup (\textit{top}).
    }
    

    \label{fig:overview}
\end{figure}

Our proposed system, depicted in \Cref{fig:overview}, consists of, firstly, a teleoperation interface.
Like previous research work~\cite{si2021review}, we demonstrate the tasks by directly teleoperating the robot. This paper presents a teleoperation system based on Virtual Reality (VR) controllers, as it has been shown to be a more intuitive interface for users to tele-operate robots~\cite{rakita2017vrteleop,whitney2019comparing}. Although similar systems have been proposed before~\cite{zhang2018vrteleop,lipton2017vrteleop,rosen2018testing} we aim to provide an inexpensive haptic feedback interface to the operator through the vibration mechanism of the controllers while safely controlling the rigid robots via compliance control.


Secondly, inspired by \cite{zhao2023learning}, we propose a method to learn variable \textit{\underline{Comp}liance Control via \underline{A}ction \underline{C}hunking with \underline{T}ransformers} (Comp-ACT). It starts by gathering demonstrations, including the Cartesian trajectory of the robots' end-effector, the measured F/T on each robot, the compliance control parameters during teleoperation (i.e., stiffness), and camera images from multiple points of view. Following that, we train a Comp-ACT policy to predict a chunk of future actions given the current state of the robot. The main differences with \cite{zhao2023learning} are that we learn a policy using the Cartesian task space instead of joint space; the external F/T is also used to condition the future action; and the stiffness parameters of the compliance controller are simultaneously learned.
In summary, we make the following contributions;
\begin{itemize}
    \item An open-source variable compliance control-based teleoperation system with haptic feedback that uses Virtual Reality (VR) controllers. This teleoperation system is particularly useful for position-controlled robots but can be easily used with other types of robots.
    \item Comp-ACT: A novel approach to learning variable compliance control from a few demonstrations via \textit{Action Chunking with Transformers}. Comp-ACT learns the reference Cartesian trajectory and the corresponding time-varying stiffness parameters simultaneously.
\end{itemize}

Our proposed method has been validated on various complex contact-rich tasks on a single arm and a bimanual robot system in simulation and real hardware.

\section{Related Work}

\subsection{Learning variable compliance control}
For robots to safely interact with the physical world, they must carefully control the contact interactions between their bodies and the objects around them. Compliance control methods allow robots to realize contact-rich manipulation tasks but require fine-tuning additional parameters to achieve safe and optimal desired behaviors. Data-driven methods have been proposed to learn such compliance parameters~\cite{abu2020variable}. In this work, we focus on rigid robots, i.e., robots that can only be actuated by the end-user via position/velocity controls. 

Reinforcement Learning (RL) is a common approach that has been used to solve tasks such as assembly~\cite{kuan1998reinforcement, beltran2020variable, ren2018learning, liu2023knowledge} and human-robot co-manipulation~\cite{dimeas2015reinforcement}. However, RL has several disadvantages, such as safety, sample efficiency, and expert knowledge requirements. RL is generally more dangerous to apply directly to real robots as its principle of learning from trial-and-error and random exploration can lead to unexpected and undesired behavior. Additionally, RL requires a lot of data from interactions with the environment and expert knowledge to craft the reward function. On the other hand, LfD is safer as it can learn from real-world data, is more sample-efficient, and does not require expert knowledge~\cite{correia2023survey,zhu2018robot}.

In this work, we use LfD to teach the robots complex contact-rich manipulation tasks. One of the challenges of LfD is the compounding error problem~\cite{asadi2018lipschitz}, where small errors in the policy's prediction lead to states further away from the demonstrated behavior. Action Chunking with Transformers (ACT)~\cite{zhao2023learning}, as its name indicated, proposed predicting a chunk of future actions at each time step instead of only predicting the immediate next action. This approach reduces the accumulation of errors. We adopted ACT and adapted it to work in task space and with force-torque information to predict the chunk of future actions that includes not only the robot's motion trajectory but also the compliance controller's stiffness parameters. We show that our proposed method is safer for rigid robots as it leads to lower contact forces with the environment.

Most similar to our method is \cite{buamanee2024bi}, which also adopts the ACT approach but adds force information and force control to achieve more robust and responsive control. However, their approach assumes that the robot can be actuated via joint torque commands. In contrast, our work focuses on rigid robots whose lowest-level actuation interface is position/velocity controls, so it is impossible for end-users to actuate them through torque commands.

\subsection{Teleoperation Systems}

Teleoperation provides the most direct method for information transfer within demonstration learning~\cite{si2021review}. 
Several teleoperation interfaces have been proposed based on different teaching devices, such as the 3D mouse, the joystick~\cite{chen2003programing}, the VR controller~\cite{rakita2017vrteleop,zhang2018vrteleop,lipton2017vrteleop,martinez2015telepresence,peppoloni2015vrtelepresense,su2022mixed}, vision-based teleoperation \cite{wang2020motion,kobayashi2023lfdt} and twin systems for leader-follower control \cite{saigusa2022imitation,zhao2023learning,wu2023gello}. 
Compared to the 3D mouse or the joystick, VR controller-based teleoperation systems have been shown to be more intuitive for novice users, as the user can direct the robot using the natural space of his/her own hand~\cite{rakita2017vrteleop,whitney2019comparing}. Furthermore, these systems do not require custom-designed hardware compared to twin systems like ~\cite{wu2023gello}; instead, they use commercially available VR hardware.

Most of the previously proposed VR-based teleoperation systems implemented a virtual world where the human operator is aligned with the simulated robot. For simplicity, this work uses only the VR controllers' tracking capability. 
Authors in \cite{GiammarinoGA23} proposed a tele-impedance framework using a custom-designed interface similar to VR controller, validated on a torque-controlled robot. Our teleoperation interface, on the other hand, adopts a commercially available VR controller combined with a compliance controller, which is required for adopting the system for rigid robots.
Additionally, we use haptic feedback as it has been shown to improve robotic arm control~\cite{motamedi2016haptic}. Our haptic feedback interface uses the VR controller's vibration mechanism, similar to~\cite{mandlekar2018roboturk}, where they use the phone vibrations. 


\section{Control System}
\subsection{Teleoperation System} \label{subsec:teleoperation}
\begin{figure}[t]
    \centering
    \includegraphics[width=0.95\linewidth]{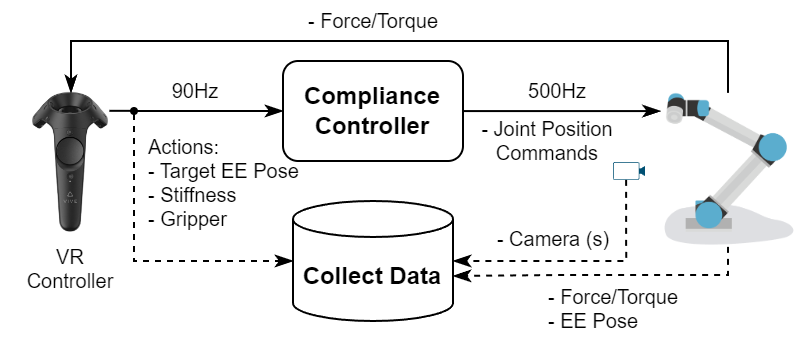}
    \caption{Teleoperation system for data collection using VR controllers. The controller receives the contact force information from the robot's wrist F/T sensor and uses it to provide haptic feedback to the operator through the rumbling (vibration). The same concept is applied to bimanual tasks where a different VR controller is connected to each arm.}
    \label{fig:teleop-system}
\end{figure}

    An overview of our teleoperation system is described in \Cref{fig:teleop-system}. We developed a teleoperation system based on Virtual Reality (VR) controllers, specifically the HTC Vive Cosmos Elite. 
    
    The VR controller enables the user to control the robot by providing the reference position for the robot to follow, the commands to the gripper, and, optionally, the desired degree of compliance. For safety reasons, a rigid robot should not be teleoperated to solely follow a desired trajectory. By design, rigid robots consider any environmental resistance a position error. Thus, they try to correct the error by exerting large torques, which may damage the environment or the robot. For this reason, our teleoperation system uses a Cartesian compliance control approach to allow the operator to demonstrate contact-rich manipulation tasks safely. Additionally, through our teleoperation interface, the operator can receive haptic feedback through the vibration mechanism of the controllers. 
    During teleoperation, the user can operate the robot more safely by considering the haptic feedback and adapting its behavior accordingly. For example, the user could move away from the contact or change the controller's degree of compliance (stiffness).
    
\begin{figure}[t]
    \centering
    \includegraphics[width=0.75\linewidth]{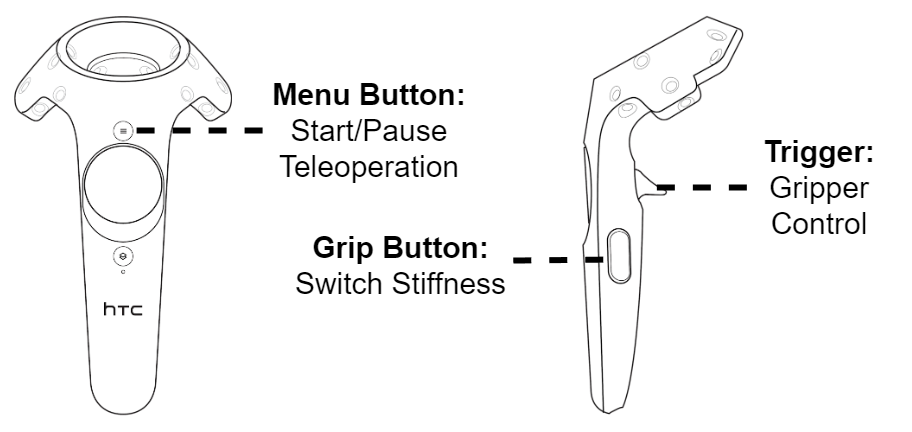}
    \caption{Button bindings for HTC Vive controller on teleoperation task.}
    \label{fig:vr_controller}
\end{figure}

    Our teleoperation system works as follows. First, we use PyOpenVR~\cite{pyopenvr}, a Python wrapper for ValveSoftware's OpenVR library, to obtain the tracked pose of the controllers at 90Hz. We are interested in the relative translation and orientation of the controller from a defined starting pose. The relative motion of the controller is then mapped to the pose of the robot's end-effector (EE) and defined as the target pose for the controller to follow. The operator sets the starting pose of the controller used to compute the relative motion through the \textit{menu button} of the controller, which also enables/disables the teleoperation.  This control mechanism allows the user to operate the robot easily regardless of the workspace size difference due to the different lengths of the human arm and the robot arm. Additionally, the \textit{trigger} on the controller is used to signal the desired gripper pose, which ranges from zero to a hundred percent of the maximum opening width of the gripper. The \textit{grip button} switches between different degrees of compliance, i.e., the compliance controller stiffness. The haptic feedback to the user is achieved by mapping the contact force measure at the robot's wrist F/T sensor to the level of intensity of the controller's vibration. The button bindings are shown in \Cref{fig:vr_controller}.

\subsection{Cartesian Compliance Control} \label{sec:controller}
    The Forward Dynamics Compliance Control (FDCC) method \cite{scherzinger2017forward} is used to control the robots with commands from either the teleoperation system or the learned policy. FDCC combines three control principles, Impedance Control, Admittance Control, and Force Control, into one new strategy to realize Cartesian compliance control. As a critical component in FDCC, forward dynamics simulations of a virtual model are leveraged to map Cartesian inputs directly to joint control commands, leading to good stability in singularities. 
    Our proposed system uses the open-source implementation~\cite{cartesian_controllers}
    of FDCC for ROS.

    In this work, we consider the following parameters as controllable and time-variant: the goal Cartesian position of the robot's EE $x_{g}$ and the stiffness of the controller $K$. Other parameters, including the goal wrench $F_{g}$, were manually fine-tuned for the teleoperation and kept constant for all tasks and experiments. For the interested reader, see more details of FDCC's parameters in \cite{scherzinger2017forward}.

\begin{figure*}[tbp]
    \centering
    \includegraphics[width=0.9\textwidth]{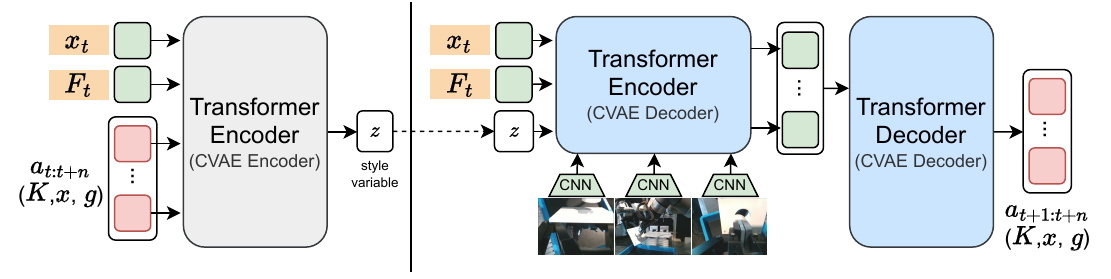}
    \caption{Comp-ACT's Network architecture. \textit{Left:} The action sequence, consisting of $n$ robot states (stiffness $K$, target EE's Cartesian pose $x$, and gripper pose $g$), are encoded alongside the current Cartesian pose $x_t$ and the measured F/T $F_t$ by the CVAE encoder. This network is discarded at inference time. \textit{Right:} The policy inputs are images from multiple viewpoints, the current Cartesian pose, and the measured F/T. The policy predicts a sequence of $n$ future actions.}
    \label{fig:act}
\end{figure*}

\section{Comp-ACT: Learning Variable Compliance Control From a Few Demonstrations}

We aim to learn Cartesian variable compliance control from a few demonstrations with the aid of the Transformer model.
To achieve this, we propose \textit{Comp-ACT}, which modifies ACT to enable it to handle Cartesian compliance control. Comp-ACT predicts target Cartesian EE poses and robot stiffness parameters conditioned on the current observation, leveraging the Transformer's capability to predict multimodal action data. 
The predicted desired Cartesian pose and stiffness parameters serve as inputs for the compliance controller.
This enables the robots to adjust their arm stiffness autonomously depending on the task phase while following the desired motion trajectory to execute the task.
\Cref{fig:act} shows the architecture of Comp-ACT.

\subsection{Transformer-based Behavioral Cloning}
The Transformer-based policy has shown great promise in processing long-horizon multimodal action data in an autoregressive manner. 
We built Comp-ACT upon Action Chunking with Transformers~\cite{zhao2023learning} to learn Cartesian EE trajectory and robots' stiffness from demonstrations. ACT predicts a sequence of future actions instead of a single action like in standard Behavioral Cloning (BC), mitigating compounding errors. Following ACT, we learn the policy as the decoder of a conditional variational autoencoder (CVAE). It predicts a sequence of action conditioned on the current observation, including RGB images, F/T sensor readings, and proprioception. \Cref{fig:act} illustrates the architecture of the CVAE. The predicted Cartesian EE pose and stiffness are then fed into the compliance controller described in \Cref{sec:controller}. 
Each robot's action includes three components: a 6D Cartesian pose (with position and orientation in 3D space, the latter in axis-angle form), a 1D gripper action (non-binary desired opening width), and a 12D vector to represent the stiffness parameters. Consequently, the bimanual robot's action space is 38-dimensional. We use the Cholesky representation to encode the stiffness matrix for translation and rotation for better extensibility to full stiffness matrix, similar to~\cite{abu2018force}. The Cholesky representation is the flattened form of the upper triangular matrix $\hat{K}$ from the Cholesky decomposition of the stiffness matrix $K$, where $K = \hat{K}^T \hat{K}$.

\begin{table}[t]
\centering
\setlength{\extrarowheight}{0pt}
\addtolength{\extrarowheight}{\aboverulesep}
\addtolength{\extrarowheight}{\belowrulesep}
\setlength{\aboverulesep}{0pt}
\setlength{\belowrulesep}{0pt}
\caption{EXPERIMENTAL CONDITIONS}
\label{table:experimental_conditions}
\begin{tabular}{lcccl} 
\toprule
\multicolumn{1}{c}{Task} & Type & \begin{tabular}[c]{@{}c@{}}\# of \\Demos\end{tabular} & \begin{tabular}[c]{@{}c@{}}\# of\\ Cam\end{tabular} & \multicolumn{1}{c}{\begin{tabular}[c]{@{}c@{}}Stiffness Mode\\ Transition\end{tabular}} \\ 
\midrule
\rowcolor[rgb]{0.875,0.875,0.875} \begin{tabular}[c]{@{}>{\cellcolor[rgb]{0.875,0.875,0.875}}l@{}}\textbf{[Simulation]}\\\textbf{Bimanual Wiping}\end{tabular} & Bimanual & 30 & 2 & \begin{tabular}[c]{@{}>{\cellcolor[rgb]{0.875,0.875,0.875}}l@{}}(L) Mid \\(R) Mid$\rightarrow$Low\end{tabular} \\
\begin{tabular}[c]{@{}l@{}}\textbf{Picking \& }\\\textbf{Insertion}\end{tabular} & Single & 30 & 2 & Mid$\rightarrow$Low \\
\rowcolor[rgb]{0.875,0.875,0.875} \textbf{Wiping} & Single & 20 & 2 & Mid$\rightarrow$Low \\
\textbf{Drawing} & Single & 20 & 2 & Mid$\rightarrow$Low \\
\rowcolor[rgb]{0.875,0.875,0.875} \begin{tabular}[c]{@{}>{\cellcolor[rgb]{0.875,0.875,0.875}}l@{}}\textbf{Peg-in-Hole }\\\textbf{Cylinder}\end{tabular} & Bimanual & 20 & 3 & \begin{tabular}[c]{@{}>{\cellcolor[rgb]{0.875,0.875,0.875}}l@{}}(L) Mid$\rightarrow$High \\(R) Mid$\rightarrow$Low\end{tabular} \\
\begin{tabular}[c]{@{}l@{}}\textbf{Peg-in-Hole }\\\textbf{Cuboid}\end{tabular} & Bimanual & 30 & 3 & \begin{tabular}[c]{@{}l@{}}(L) Mid$\rightarrow$High\\ (R) Mid$\rightarrow$Low\end{tabular} \\
\bottomrule
\end{tabular}
\end{table}

\subsection{Data Collection}
The demonstration data was collected using the teleoperation system described in \Cref{subsec:teleoperation} and the compliance controller described in \Cref{sec:controller}. The action to be saved is taken from the VR controller, consisting of the target EE pose, the gripper's command, and the desired degree of compliance.
This work considered three predefined stiffness modes during the compliant teleoperation of robotic arms. Specifically, the stiffness values considered are 250, 500, and 750 for low, medium, and high, respectively. The same stiffness is applied to all diagonal values of the stiffness matrix. During demonstrations, the operator can switch between two selected stiffness modes out of three using the \textit{grip button} on the VR controller. For each task and arm, one mode is always set to medium stiffness, while the other alternates between high and low stiffness, predetermined based on the task requirements. \Cref{table:experimental_conditions} indicates the stiffness modes available on each task.

\section{Experiments}
Both simulation and real-world experiments were conducted to evaluate Comp-ACT's performance. We rely on simulation to compare Comp-ACT against ACT without compliance control due to the risks of implementing ACT on rigid, position-controlled robots in the real world. Real-world experiments assessed the proposed Comp-ACT's performance in challenging tasks. Two of the co-authors performed demonstrations to provide a diversity of motion strategies for each task.

\subsection{Experimental Setup}
\begin{figure*}[t]
    \centering
      \begin{@twocolumnfalse}
        {
        \vspace{0.25cm}
        \normalsize TASK DESCRIPTION} 
        \vspace{0.25cm}
      \end{@twocolumnfalse}
    
    \includegraphics[width=0.95\textwidth]{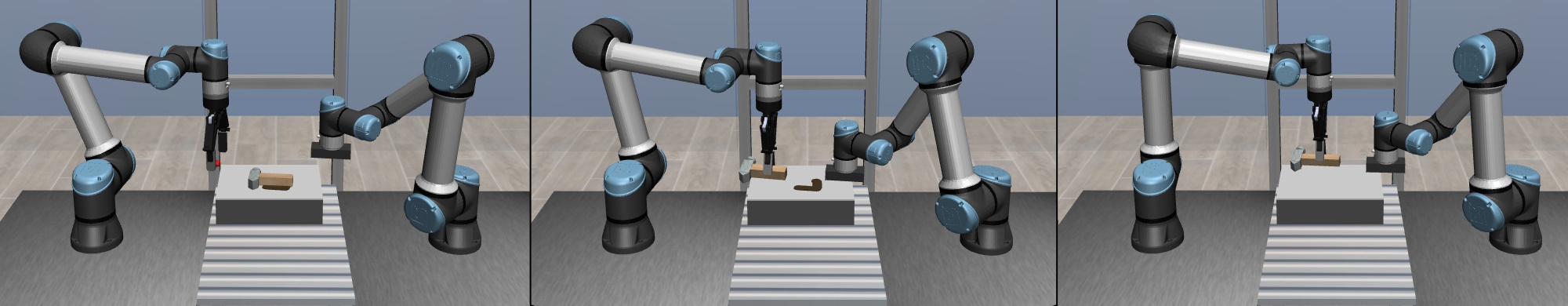}
    \caption{\textbf{Bimanual Wiping task in \textit{robosuite}}: The goal is to pick up the hammer (left), wipe the exposed dirt beneath the hammer with the other arm (center), and then return the hammer to its initial place (right). During demonstrations, The left arm is kept at a constant medium stiffness. In contrast, the right arm begins with a medium stiffness mode to approach the table, then switches to a low stiffness mode to delicately apply force against the table during the wiping motion. The task is randomized by rotating the hammer between $\pm 15\degree$, and changing the shape of the dirt. 30 demonstrations were collected for this task.}
    \label{fig:bimanual_wiping}
    
    \includegraphics[width=0.95\textwidth]{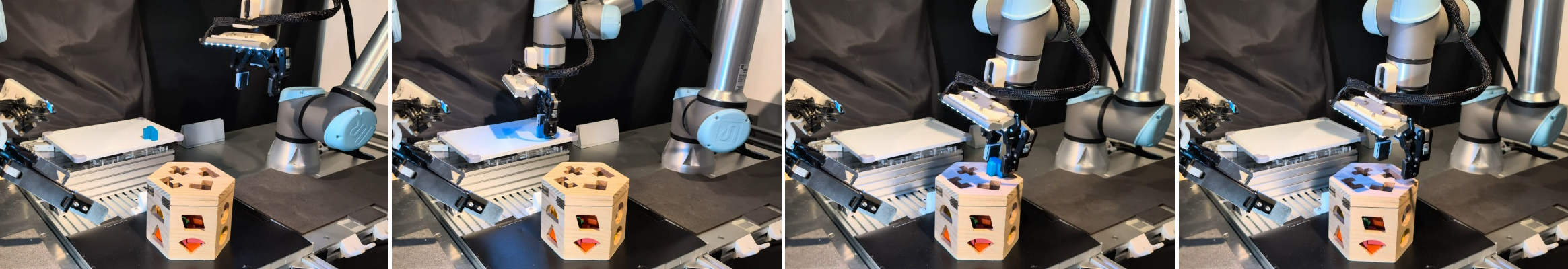}
    \caption{\textbf{Single-arm Picking \& Insertion}: 
    The task consists of picking up a toy wooden peg and inserting it into its corresponding hole in a wooden box. A proper grasp is required to facilitate the alignment of the peg with the hole. During demonstrations, the demonstrator utilizes a medium stiffness mode for general manipulation and switches to a low stiffness mode for the insertion phase that involves physical contact. The peg is placed at the edge of the whiteboard with a random rotation of about $\pm 15\degree$. The wooden box's position is not fixed, so the robot can move it during the peg insertion. 30 demonstrations were collected for this task.
    }
    \label{fig:single_arm_picking}
    
    \includegraphics[width=0.95\textwidth]{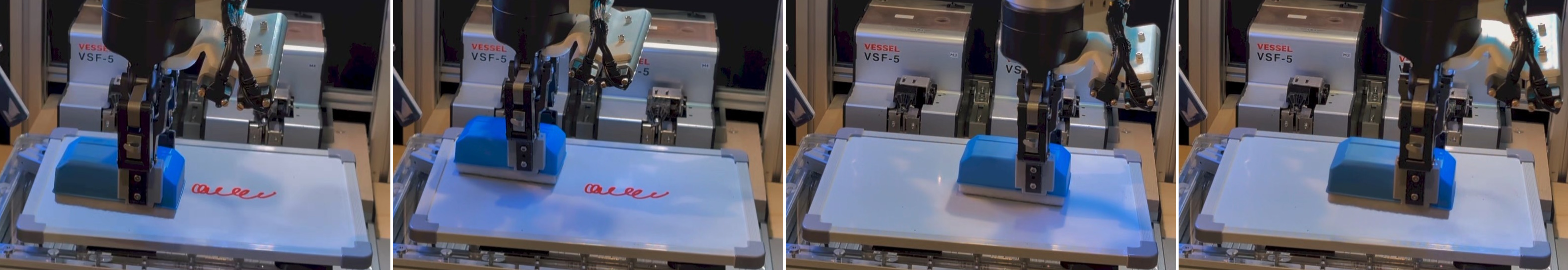}
    \caption{\textbf{Single-arm Wiping}: The goal is to wipe marks on a whiteboard. The task starts with the robot already holding the eraser. 
    During demonstrations, the demonstrator begins with a medium stiffness mode to approach and then switches to a low stiffness mode to apply force to the whiteboard. The task requires applying sufficient contact force to the board during the wiping motion. The task starts with a random mark drawn in an area of $\pm 5$~cm from the center of the board. 20 demonstrations were collected for this task.}
    \label{fig:single_arm_writing}
    
    \includegraphics[width=0.95\textwidth]{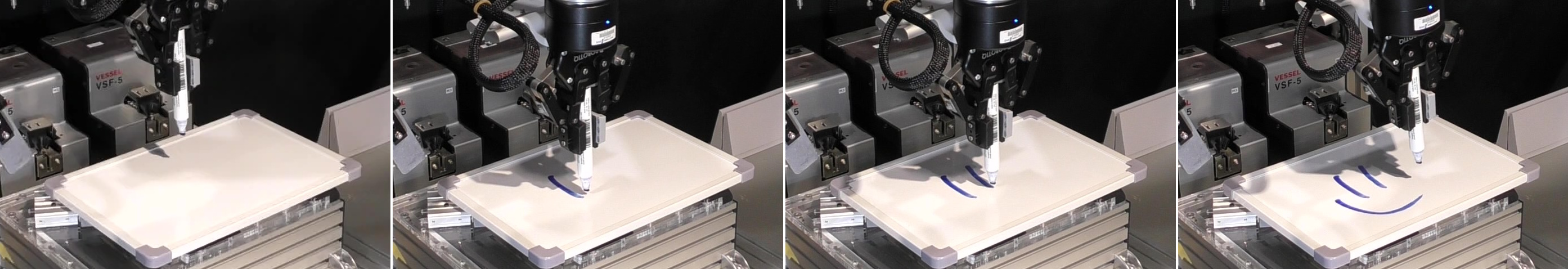}
    \caption{\textbf{Single-arm Drawing}: The goal is to draw a simple emoji smile on a whiteboard. The task starts with the robot already grasping a pen. The robot must write three parts of a face while keeping a consistent force while in contact. This task is interesting in that it requires frequent switching between in-contact writing and free motion. 
    It also requires careful coordination of the positional relationship between each part of the smile. During demonstrations, the demonstrator employs a medium stiffness mode for the free motion between drawing stages and transitions to a low stiffness mode to apply force during the in-contact drawing. 20 demonstrations were collected for this task.}
    \label{fig:single_arm_wiping}
    
\end{figure*}

We used two UR5e robot arms with built-in force-torque sensors in simulation and real-world experiments, with bimanual and single-arm tasks. Two cameras, Realsense SR305, are attached to the wrist of each arm. Another static camera with a wider field of view is in front of the task space. Different combinations of these cameras were used as observations on each task and described in \Cref{sec:tasks}. \Cref{fig:overview} shows our experimental setup. 
For simulation, we use the physics engine MuJoCo~\cite{todorov2012mujoco}. A virtual twin of our robotic setup was created as a new \textit{robosuite}~\cite{robosuite2020} environment, as shown in \Cref{fig:bimanual_wiping}. 
The robotic system runs on a single machine with a CPU Intel i7-12700F and a GPU NVIDIA GeForce RTX 3070. The simulation and the policy training are performed on a different machine with a GPU NVIDIA GeForce RTX 4090.

\begin{figure*}[t]
    \centering
    \includegraphics[width=0.95\textwidth]{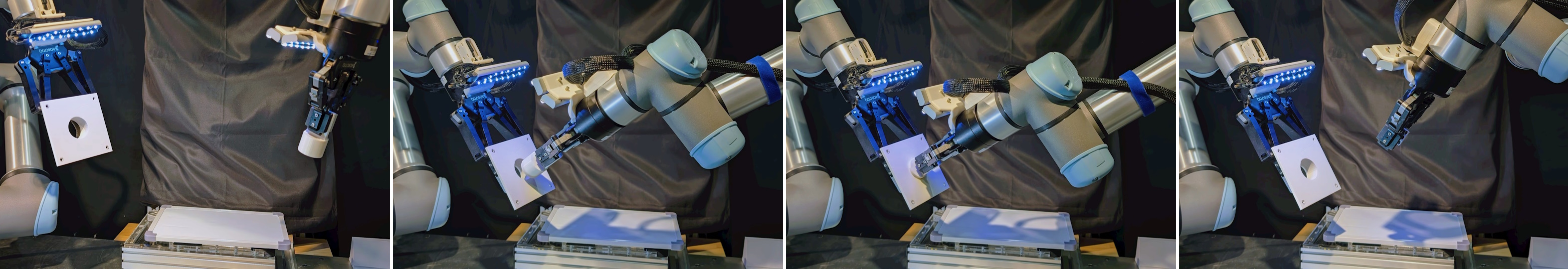}
    \caption{\textbf{Bimanual Insertion (Cylinder)}: A 3D-printed peg and matching hole with tolerance of $2$~mm. The relative position and orientation of the two arms need to be accurately aligned to complete the round-shaped peg insertion. During demonstrations, both arms start with medium stiffness mode during contact-free motion. The left arm switches to high stiffness to resist motion during the insertion. On the contrary, the right arm switches to low stiffness to slow the robot and allow force to guide the insertion. 20 demonstrations were collected for this task.}
    \label{fig:bimanual_insertion_cylinder}
    
    \includegraphics[width=0.95\textwidth]{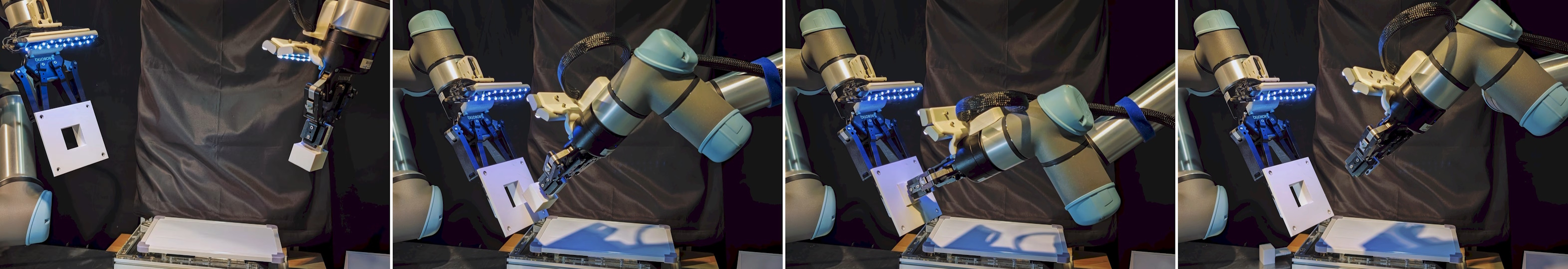}
    \caption{\textbf{Bimanual Insertion (Cuboid)}: A 3D-printed peg and matching hole with tolerance of $2$~mm. Besides matching the peg and hole's surface position and orientation, the cuboid insertion also requires aligning each corner by rotating around the direction perpendicular to the surface. The same stiffness switching strategy described in \Cref{fig:bimanual_insertion_cylinder} was used during demonstrations. 30 demonstrations were collected for this task.}
    \label{fig:bimanual_insertion_prism}

\end{figure*}


\subsection{Tasks}
\label{sec:tasks}
Comp-ACT was evaluated with 6 challenging contact-rich manipulation tasks, one in simulation and five in our real-world robotic system. All tasks are described in Figures \ref{fig:bimanual_wiping} to \ref{fig:bimanual_insertion_prism}.
The summary of experimental conditions for each task is shown in the \Cref{table:experimental_conditions}. 

For every task, a policy was learned from scratch for 20000 epochs, corresponding to 2-5 hours in the actual time, depending on the number of demonstrations and the number of cameras used.

\section{Results}
\subsection{Simulation Task}
We evaluated our proposed method Comp-ACT compared to ACT~\cite{zhao2023learning} on simulation with the bimanual wiping task described in \Cref{fig:bimanual_wiping}. The effect of adding F/T data as observation was also investigated. The same demonstrations were used to compare both methods fairly. For ACT, the action was defined as the joint positions of each robot, and we confirmed that replaying the demonstrations using joint-position control achieves similar contact forces. The learned policies were tested 10 times each. 
\subsubsection{Comp-ACT vs ACT} 
\begin{figure}[t]
    \centering
    \includegraphics[width=\linewidth]{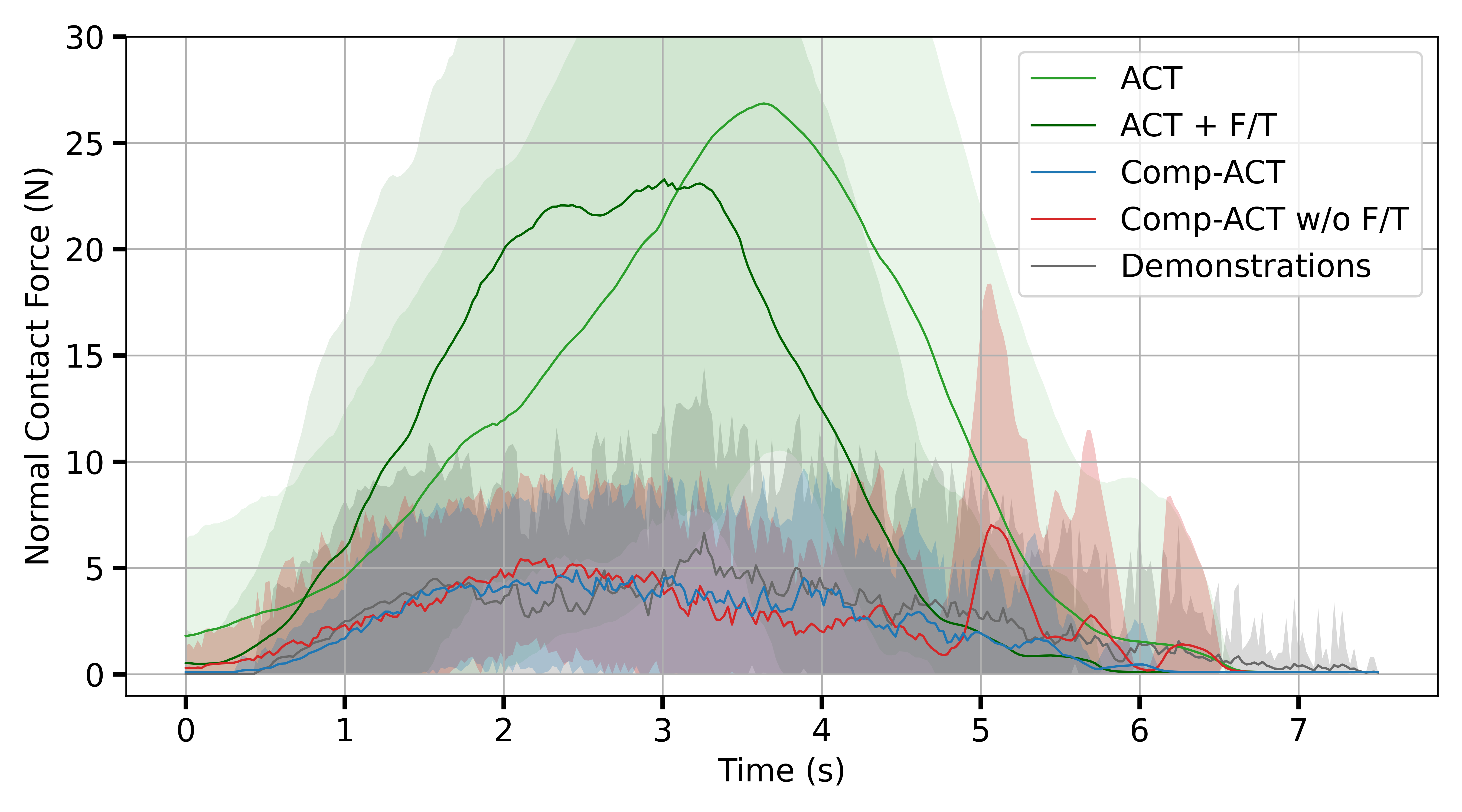}
    \caption{Force comparison for the simulated bimanual wiping task. Only the contact force measured by the wiping robot is considered. The average contact force (continuous line) and standard deviation (shaded colored area) of 20 trials for each policy and the demonstrations are shown.}
    \label{fig:result_sim_wiping_ft}
\end{figure}

Both policies learned and solved the task correctly, i.e., the hammer is picked and lifted, the mark on the table wiped, and then the hammer is returned to its original place. However, as shown in \Cref{fig:result_sim_wiping_ft}, when we examined the contact force applied to the table, it is clear that when using joint-position control with the ACT policy, the robot would apply force to the table of more than 5 times higher than when using our proposed Comp-ACT method. In the case of ACT, minor errors in the trajectory predicted by the policy would result in significant forces being applied to the table due to the joint-position control scheme. In contrast, with our compliance control scheme, minor errors in the trajectory would not result in significant forces due to the low stiffness used while wiping.
\subsubsection{F/T as observation vs Not} 
Adding F/T data to the observation of ACT policy (\textit{ACT + F/T}) helped suppress the applied force compared to ACT without F/T observation (\textit{ACT}), yet both applied excessive force to the table. However, there was no significant difference in the force plot between Comp-ACT and Comp-ACT without F/T. We argue that it is because simply following the predicted reference Cartesian trajectory with a specified stiffness should already achieve relatively lower contact force similar to the demonstration, thus the F/T data did not have much information to the policy for this wiping task. 
\begin{figure}[t]
    \centering
    \includegraphics[width=\linewidth]{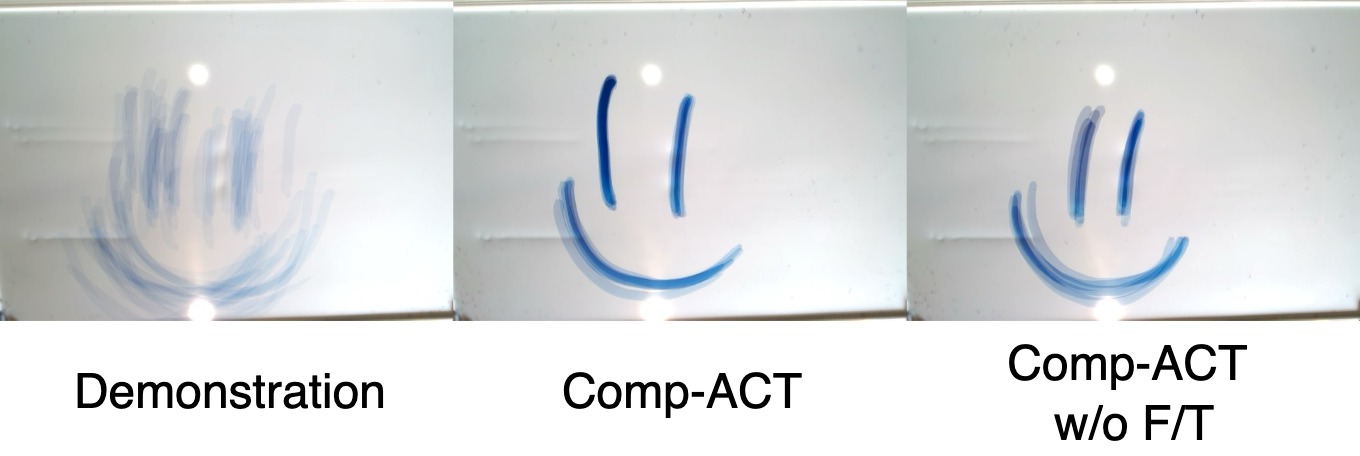}
    \caption{Comparison of drawings. Averaged images from demonstrations (\textit{Left}), Comp-ACT (\textit{Middle}), and Comp-ACT without F/T data (\textit{Right}).} \textit{}
    \label{fig:drawing_results}
\end{figure}
\renewcommand{\arraystretch}{1.5}
\begin{table}[t]
\centering
\caption{REAL-WORLD TASKS SUCCESS RATE
}
\label{table:real-results}

\begin{tabular}{|c|c|c|c|} 
\hline
\multicolumn{2}{|c|}{\multirow{2}{*}{Task Name}} & \multicolumn{2}{c|}{Comp-ACT} \\ 
\cline{3-4}
\multicolumn{2}{|c|}{} & F/T & w/o F/T \\ 
\hline
\multirow{3}{*}{\begin{tabular}[c]{@{}c@{}}Single \\ Robot\end{tabular}} & Picking \& Insertion & \textbf{70} & 35 \\ 
\cline{2-4}
 & Wiping & 70 & \textbf{100} \\ 
\cline{2-4}
 & Drawing & \textbf{100} & \textbf{100} \\ 
\hline
\multirow{2}{*}{Bimanual} & Peg-in-Hole (Cylinder) & \textbf{100} & 40 \\ 
\cline{2-4}
 & Peg-in-Hole (Cuboid) & \textbf{70} & 50 \\
\hline
\end{tabular}
\end{table}

\subsection{Real-World Tasks}
In the five real-world tasks, we evaluated the efficacy of Comp-ACT in complex, real-world, contact-rich manipulation tasks. As is also done in the simulation task, we conducted an ablation study to determine the effectiveness of F/T data as observational input for the policy. The results are summarized in \Cref{table:real-results} and discussed below:

\subsubsection*{Picking \& Insertion}
This task is challenging because a successful insertion depends greatly on the orientation of the peg after picking. Comp-ACT policies, both with and without F/T, consistently succeeded in picking the peg. The success rates for the insertion phase were 70\% with F/T and 35\% without, indicating the effectiveness of F/T data in enhancing peg-in-hole task performance. 



\subsubsection*{Wiping}
For this task, a trial was considered successful if most marks were erased in a single wiping motion. The failure cases resulted from the policy ending its wiping motion too early. Only in this task did we observe the Comp-ACT policy without F/T data performing better than having the additional F/T information. This result indicates the task-dependent importance of the F/T data. 
The F/T data may have erroneously signaled the policy to stop the wiping motion and return to the initial position. In contrast, the policy Comp-ACT without F/T data performed longer wiping motions, even longer than those observed in the demonstrations.

\subsubsection*{Drawing}
\begin{figure}[t]
    \centering
    \includegraphics[width=0.95\linewidth]{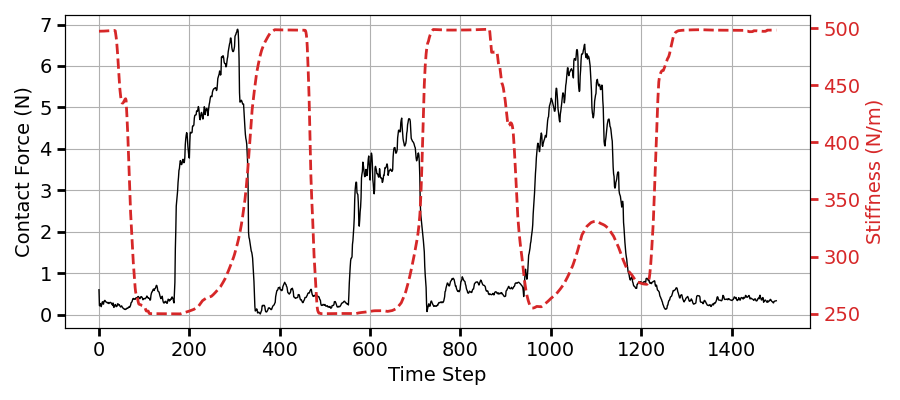}
    \caption{Drawing task: contact force versus predicted stiffness action during policy evaluation.}
    \label{fig:stiffness_ft_plot}
\end{figure}

In this case, success was defined subjectively by the resemblance of the drawing to an emoji smile. \Cref{fig:drawing_results} shows the drawings by demonstrators and trained policies (with and without F/T). All trials successfully produced emoji smiles with accurately placed eyes and mouths, using appropriate force for clear, continuous lines. Quantitative analysis of similarity to the demonstrated smiles using the Structural Similarity Index (SSIM)~\cite{wang2004ssim} showed negligible differences in performance between Comp-ACT with an average similarity score of 84.2\% and Comp-ACT without F/T at 84.3\%. Similarly to the wiping task, policies without F/T data exhibited longer motions.
Furthermore, the policy's predicted stiffness action versus the normal force applied during the drawing task is visualized in \Cref{fig:stiffness_ft_plot}. The policy successfully chooses lower stiffness while in contact with the whiteboard, which realizes avoiding excessive force.

\subsubsection*{Peg-in-Hole tasks}
The two bimanual peg-in-hole insertion tasks are the most challenging of the ones considered in this paper. While the robot's initial position remained the same in each demonstration, the pose where the robots joined to perform the insertion changed every time, as the demonstrators could not reproduce the same motions perfectly. Nevertheless, we observed high success rates for both tasks with our proposed method: 100\% and 70\% for the cylindrical peg insertion and the cuboid peg insertion, respectively. The success rate drops significantly when the F/T data is not included in the Comp-ACT policy: 40\% and 50\% for cylindrical and cuboid peg insertion, respectively. These results further corroborate the importance of F/T data for peg insertion tasks. In both tasks, the failure cases observed mainly consisted of misalignment of the parts and subsequently applying excessive contact force or releasing the peg too early before the insertion. Videos of the policy rollouts including the failure cases can be found in our \hyperlink{https://github.com/omron-sinicx/CompACT}{project website}.

%

\section{Limitations and Future Work}
Although Comp-ACT has shown promising results, there are limitations and potential avenues for future research. Our proposed teleoperation system enables users to define different levels of compliance that increase the safety of the robots when interacting with the environment. However, the different compliance modes have to be predefined for each task. An interesting research avenue is to estimate the desired compliance mode online based on the user's intent to make the data collection easier for the operator.
Additionally, we consider only the fine-tuning and learning of the compliance controller's stiffness parameter, as described in \Cref{sec:controller}. Adaptively learning other of the controller's parameters may increase the responsiveness and smoothness of the robots.

Furthermore, in this study, we have shown that Comp-ACT can learn complex manipulation tasks by learning a policy per task, which limits the generalization to small variations of each task. Future research will explore the feasibility of generalizing to a wider range of task variations or multiple tasks simultaneously. 


\section{Conclusion}
We studied the problem of learning contact-rich manipulation tasks from a few demonstrations. Our focus was on enabling rigid robots to tackle these types of manipulation tasks safely. To that end, we presented a novel teleoperation system with VR-based haptic feedback and an imitation learning method for learning variable compliance control called Comp-ACT. 
Our proposed system enables users to teleoperate a single or bimanual robotic system to safely collect demonstrations of challenging manipulation tasks. 
Our evaluation showed that Comp-ACT could learn complex tasks with a high success rate from 20 to 30 demonstrations.

Additionally, we showed that the baseline ACT method~\cite{zhao2023learning}, with its joint position-based control, would not be suitable for rigid robots as errors in the predicted trajectory could result in undesired large contact forces that may damage the manipulated objects or the robot. On the contrary, Comp-ACT reduces the risk of excessive forces by learning the stiffness parameters of the underlying compliance controller, enabling safe task execution with variable compliance in Cartesian space. 

\bibliographystyle{IEEEtran}
\bibliography{root}

\end{document}